\def\year{2021}\relax
\documentclass[letterpaper]{article} % DO NOT CHANGE THIS
\usepackage{aaai21}  % DO NOT CHANGE THIS
\usepackage{times}  % DO NOT CHANGE THIS
\usepackage{helvet} % DO NOT CHANGE THIS
\usepackage{courier}  % DO NOT CHANGE THIS
\usepackage[hyphens]{url}  % DO NOT CHANGE THIS
\usepackage{graphicx} % DO NOT CHANGE THIS
\usepackage{natbib}  % DO NOT CHANGE THIS AND DO NOT ADD ANY OPTIONS TO IT
\usepackage{caption} % DO NOT CHANGE THIS AND DO NOT ADD ANY OPTIONS TO IT
\usepackage{graphicx} 
\usepackage{subfigure}
\usepackage{booktabs}
\usepackage{multirow}
\usepackage[namelimits]{amsmath} 
\usepackage{amssymb}           
\usepackage{amsfonts}            
\usepackage{mathrsfs}  
\usepackage{amsthm,amssymb}
\usepackage{color}
\usepackage{xcolor}%定义了一些颜色  
\usepackage{colortbl,booktabs}%第二个包定义了几个*rule 
\usepackage{fancybox}
\usepackage{framed} 
\usepackage{tikz}
\usepackage{pgfplots}
\usepackage{array}
\usepackage{microtype} 
\usepackage{tcolorbox} 
\usepackage[tikz]{bclogo}
\usepackage{lipsum} 
\usepackage{bbding}
\usepackage{pifont}
\usepackage{wasysym}
\usepackage{amssymb}
\title{Learning Universal Representations from Word to Sentence}
\author{
 Yian Li\textsuperscript{\rm 1,2,3},
 Hai Zhao\textsuperscript{\rm 1,2,3,
 \thanks{Corresponding author. This paper was partially supported by National Key Research and Development Program of China (No. 2017YFB0304100) and Key Projects of National Natural Science Foundation of China (U1836222 and 61733011).}}\\
}
 \affiliations{
 \textsuperscript{\rm 1}Department of Computer Science and Engineering, Shanghai Jiao Tong University\\
 \textsuperscript{\rm 2}Key Laboratory of Shanghai Education Commission for Intelligent Interaction\\
 and Cognitive Engineering, Shanghai Jiao Tong University, Shanghai, China\\
 \textsuperscript{\rm 3}MoE Key Lab of Artificial Intelligence, AI Institute, Shanghai Jiao Tong University\\
 {\tt liya19@sjtu.edu.cn, zhaohai@cs.sjtu.edu.cn} \\
}
\begin{document}

\maketitle

\begin{abstract}
Despite the well-developed cut-edge representation learning for language, most language representation models usually focus on specific level of linguistic unit, which cause great inconvenience when being confronted with handling multiple layers of linguistic objects in a unified way. Thus this work introduces and explores the universal representation learning, i.e.,  embeddings of different levels of linguistic unit in a uniform vector space through a task-independent evaluation. We present our approach of constructing analogy datasets in terms of words, phrases and sentences and experiment with multiple representation models to examine geometric properties of the learned vector space. Then we empirically verify that well pre-trained Transformer models incorporated with appropriate training settings may effectively yield universal representation. Especially, our implementation of fine-tuning ALBERT on NLI and PPDB datasets achieves the highest accuracy on analogy tasks in different language levels. Further experiments on the insurance FAQ task show effectiveness of universal representation models in real-world applications.
\end{abstract}

\section{Introduction}
Encoding linguistic units such as words, phrases or sentences into low-dimensional vectors has been the core and preliminary task for deep learning of natural language. The current language representation learning is usually done in different individual levels, typically, word or sentence. The former includes pioneering works such as word2vec, GloVe and fastText \cite{word2vec,glove,fasttext}, and the latter includes the very recent so-called contextualized representations such as ELMo, GPT, BERT, XLNet and ELECTRA \cite{elmo,gpt-1,bert,xlnet,electra}. Nevertheless, few works were done to uniformly learning and representing linguistic units in different hierarchies in the same vector space. Actually, nearly all existing work still focus on individual granular language unit for representation learning \cite{linear_1,linear_3}.

However, universal representation among different levels of linguistic units may offer a great convenience when it is needed to handle free text in language hierarchy in a unified way. As well known that, embedding representation for a certain linguistic unit (i.e., word) enables linguistics-meaningful arithmetic calculation among different vectors, also known as word analogy. For example, \emph{vector (``King") - vector (``Man") + vector (``Woman")} results in \emph{vector (``Queen")}. Thus universal representation may generalize such good analogy features or meaningful arithmetic operation onto free text with all language levels involved together. For example, \textit{Eat an onion} : \textit{Vegetable} :: \textit{Eat a pear} : \textit{Fruit}. 

In this paper, we explore the regularities of representations including words, phrases and sentences in the same vector space. To this end, we introduce universal analogy tasks derived from Google's word analogy dataset. In addition, we train a Transformer-based model and compare it with currently popular representation methods. Experimental results demonstrate that well-trained Transformer-based models are able to map sequences of variable lengths into a shared vector space where similar sequences are close to each other. Meanwhile, addition and subtraction of embeddings reflect semantic and syntactic connections between sequences. In addition, we explore the applicability of this characteristic in retrieval-based chatbots by evaluation on an insurance FAQ task, where the universal representation models significantly outperform TF-IDF and BM25. 

\section{Related Work}
\subsection{Representation Methods}
Neural language models can be simply divided into two categories from the perspective of linguistic unit types: word embeddings and sentence embeddings.

Earlier research focuses on learning high-quality word vectors. Inspired by \cite{nnlm}, \cite{word2vec} take advantage of a large corpus to train word embeddings in an unsupervised manner. \cite{glove} present GloVe by combining context window with global and local statistics. \cite{fasttext} introduce character-level $n$-gram features to enrich the meaning of word embeddings. Different from the above mentioned models, ELMo raised by \cite{elmo} learns contextualized word representations, i.e., words have different meanings according to different contexts.

To facilitate sentence-level tasks, \cite{skip-thought,quick-thought,infersent,gensen} train sentence embeddings using recurrent neural networks (RNN). With Transformer \cite{transformer} first proposed in the area of machine translation, more and more researchers turn their encoder architecture from RNN as used in ELMo to the Transformer that relies completely on attention mechanism. \cite{use} develop the Universal Sentence Encoder (USE) based on the Transformer architecture and the deep averaging network (DAN) \cite{dan}. The latest pre-trained contextualized language representations like GPT, BERT, ALBERT, RoBERTa, XLNet and ELECTRA \cite{gpt-1,gpt-2,bert,albert,roberta,xlnet,electra} are expected to handle different-sized input sentences. Fine-tuning of pre-trained BERT \cite{multidnn,sbert} further improves the performance on downstream tasks.

\subsection{Analysis on Embeddings}
Previous exploration of vector regularities mainly studies word embeddings \cite{word2vec,phrase2vec,linear_1}. After the introduction of sentence encoders and Transformer models \cite{transformer}, more works were done to investigate sentence-level embeddings. Usually the performance in downstream tasks is considered to be the measurement for model ability of representing sentences \cite{infersent,use,multidnn}. Some research proposes probing tasks to understand certain aspects of sentence embeddings \cite{probing_1,probing_2,probing_3}. Specifically, \cite{bert_embedding_1,bert_embedding_2,bert_embedding_3} look into BERT embeddings and reveal its internal working mechanisms. Besides, \cite{linear_2,linear_3} explore the regularities in sentence embeddings. Nevertheless, little work analyzes words, phrases and sentences in the same vector space. In this paper, We work on embeddings for sequences of various lengths obtained by different models in a task-independent manner. 

\subsection{FAQ Applications}
The goal of a Frequently Asked Question (FAQ) task is to retrieve the most relevant QA pairs from the pre-defined dataset given a query. Previous works focus on feature-based methods \cite{faq_feature_1,faq_feature_2,faq_feature_3}. Recently, Transformer-based representation models have made great progress in measuring query-Question or query-Answer similarities. \cite{faq_transformer} make an analysis on Transformer models and propose a neural architecture to solve the FAQ task. \cite{faq_bert_1} come up with an FAQ retrieval system that combines the characteristics of BERT and rule-based methods. In this work, we evaluate the performance of well-trained universal representation models on the FAQ task.

\section{Datasets}
As a new task, universal representation has to be evaluated in a multiple-granular analogy dataset. In this section, we introduce the procedure of constructing different levels of analogy datasets based on Google's word analogy dataset. 

\begin{table}[t]
  \centering
%   \resizebox{\textwidth}{!}{
  \begin{tabular}{|p{3cm}|p{4cm}|}\hline
     A : B :: C & Candidates\\\hline
    \textit{boy}:\textit{girl}::\textit{brother} & \textit{daughter, \textbf{sister}, wife, father, son}\\\hline
    
     \textit{bad}:\textit{worse}::\textit{big} & \textit{\textbf{bigger}, larger, smaller, biggest, better}\\\hline
     
     \textit{Beijing}:\textit{China}::\textit{Paris} & \textit{\textbf{France}, Europe, Germany, Belgium, London}\\\hline
     
     \textit{Chile}:\textit{Chilean}::\textit{China} & \textit{Japanese, \textbf{Chinese}, Russian, Korean, Ukrainian}\\\hline
  \end{tabular}
%   }
  \caption{Examples from our word analogy dataset. The correct answers are in bold}
  \label{analogy_example}
\end{table}

\subsection{Word-level analogy}
Recall that in a word analogy task \cite{word2vec}, two pairs of words that share the same type of relationship, denoted as $A$ : $B$ :: $C$ : $D$, are involved. The goal is to solve questions like ``$A$ is to $B$ as $C$ is to ?", which is to retrieve the last word from the vocabulary given the first three words. The objective can be formulated as maximizing the cosine similarity between the target word embedding and the linear combination of the given vectors:
\begin{align*}
     &d^* = \mathop{\arg\max}_{d^*} cosine(c+b-a, d)\\
     &cosine(u, v) = \frac{u\cdot v}{\|u\|\|v\|}
\end{align*}
where $a$, $b$, $c$, $d$ represent embeddings of the corresponding words and are all normalized to unit lengths.

To facilitate comparison between models with different vocabularies, we construct a closed-vocabulary analogy task based on Google's word analogy dataset through negative sampling. Concretely, for each question, we use GloVe to rank every word in the vocabulary and the top 5 results are considered to be candidate words. If GloVe fails to retrieve the correct answer, we manually add it to make sure it is included in the candidates. During evaluation, the model is expected to select the correct answer from 5 candidate words. Examples are listed in Table \ref{analogy_example}. 

 \begin{table*}[t]{}
  \centering
  \small
%   \resizebox{0.9\textwidth}{!}{
  \begin{tabular}{lccccccc}\toprule
     & capital-common&	capital-world&	city-state&	male-female	&present-participle	&positive-comparative	&positive-negative\\\midrule
    phrase-level	& \;\;6.0&	\;\;6.0&	\;\;6.0&	\;\;4.1&	4.8&	3.4&	4.4 \\
    sentence-level	&12.0&	12.0&	12.0&	10.1&	8.8	&6.1	&9.2\\\bottomrule
  \end{tabular}
%   }
  \caption{Average sequence length in phrase/sentence-level analogy datasets.}
  \label{length}
\end{table*}

\begin{table}[t]{}
  \centering
%   \small
%   \resizebox{0.9\columnwidth}{!}{
  \begin{tabular}{lrrr}\toprule
    Dataset & \#p & \#q & \#c\\\midrule
    capital-common	& 23	& 506	&5 \\
    capital-world	&116	&4524	&5 \\
    city-state	&67&	2467&	5\\
    male-female	&23	&506&	5\\
    present-participle	&33	&1056&	2\\
    positive-comparative	&37	&1322&	2\\
    positive-negative &	29	&812	&2\\\midrule
    All	&328	&11193	&- \\\bottomrule
  \end{tabular}
%   }
  \caption{Statistics of our analogy datasets. \#p and \#q are the number of pairs and questions for each category. \#c is the number of candidates for each dataset.}
  \label{statistics}
\end{table}

\subsection{Phrase/Sentence-level analogy}
To investigate the arithmetic properties of vectors for higher levels of linguistic units, we present phrase and sentence analogy tasks based on the proposed word analogy dataset. Statistics are shown in Table \ref{length} and Table \ref{statistics}.

\subsubsection{Semantic}
Semantic analogies can be divided into four subsets: ``capital-common", ``capital-world", ``city-state" and ``male-female". The first two sets can be merged into a larger dataset: ``capital-country", which contains pairs of countries and their capital cities; the third involves states and their cities; the last one contains pairs with gender relations. Considering GloVe's poor performance on word-level ``country-currency" questions ($<$32\%), we discard this subset in phrase and sentence-level analogies. Then we put words into contexts so that the resulting phrases and sentences also have linear relationships. For example, based on relationship \textit{Athens} : \textit{Greece} :: \textit{Baghdad} : \textit{Iraq}, we select phrases and sentences that contain the word ``\textit{Athens}" from the English Wikipedia Corpus\footnote{https://dumps.wikimedia.org/enwiki/latest}: ``\textit{He was hired as being professor of physics at the university of Athens.}" and create examples: ``\textit{hired by ... Athens}" : ``\textit{hired by ... Greece}" :: ``\textit{hired by ... Baghdad}" : ``\textit{hired by ... Iraq}". However, we found that such a question is identical to word-level analogy for BOW methods like averaging GloVe vectors, because they treat embeddings independently despite the content and word order. To avoid lexical overlapping between sequences, we replace certain words and phrases with their synonyms and paraphrases, e.g., ``\textit{hired by ... Athens}" : ``\textit{employed by ... Greece}" :: ``\textit{employed by ... Baghdad}" : ``\textit{hired by ... Iraq}"

\subsubsection{Syntactic}
We consider three typical syntactic analogies: Tense, Comparative and Negation, corresponding to three subsets: ``present-participle", ``positive-comparative", ``positive-negative", where the model needs to distinguish the correct answer from ``past tense", ``superlative" and ``positive", respectively. For example, given phrases ``\textit{Pigs are bright}" : ``\textit{Pigs are brighter than goats}" :: ``\textit{The train is slow}", the model need to give higher similarity score to the sentence that contains ``\textit{slower}" than the one that contains ``\textit{slowest}". Similarly, we add synonyms and synonymous phrases for each question to evaluate the model ability of learning context-aware embeddings rather than interpreting each word in the question independently. For instance, ``\textit{pleasant}" $\approx$ ``\textit{not unpleasant}" and ``\textit{unpleasant}" $\approx$ ``\textit{not pleasant}". 

\begin{table*}[t]
  \centering
%   \resizebox{\textwidth}{!}{
  \renewcommand{\multirowsetup}{\centering}
  \begin{tabular}{l|ccc|ccc|ccc|c}\toprule
  \multirow{2}{*}{\textbf{Model}}
  & \multicolumn{3}{c|}{\textbf{Word}} & \multicolumn{3}{c|}{\textbf{Phrase}} &
  \multicolumn{3}{c|}{\textbf{Sentence}}&\multirow{2}{*}{\textbf{All}} \\\cline{2-10}
  \rule{0pt}{12pt}
    & sem & syn & Avg. &sem & syn & Avg. & sem & syn & Avg. & \\\midrule
    
    GloVe \cite{glove} & 82.6 & 78.0	& 80.3	& \;\;0.0	& 40.9	& 20.5	& \;\;0.2 &	39.8 &	20.0 &	40.3  \\\midrule
    
    InferSent-1 \cite{infersent}& 69.4 &	80.5 &	75.0 &	\;\;0.0 &	59.0 &	29.5 &	\;\;0.0 &	51.1 &	25.6 &	43.4 \\
    InferSent-2& 68.8 &	88.7 &	78.8&	\;\;0.0 &	54.1 &	27.0 &	\;\;0.0 &	50.8 &	25.4 &	43.7\\
    GenSen \cite{gensen}& 44.5 &	84.4 &	64.5 &	\;\;0.0 &	54.4 &	27.2 &	\;\;0.0 &	44.9 &	22.4 &	38.0 \\
    USE-v4 \cite{use}& 73.0 &	83.1 &	78.0 &	\;\;1.8 &	63.1 &	32.5 &	\;\;0.6 &	44.1 &	22.4 &	44.3 \\
    USE-v5 & 84.2 &	86.0 &	\textbf{85.1} &	\;\;1.0 &	66.2 &	33.6 &	\;\;1.3 &	58.0 &	29.6 &	49.4 \\\midrule
    
    BERT-base \cite{bert}& 49.7 &	59.9 &	54.8 &	\;\;1.0 & 	69.3 &	35.1 &	\;\;0.2 &	68.3 &	34.2 &	41.4 \\
    BERT-large & 49.5 &	48.7 &	49.1 &	\;\;0.8 &	67.3 &	34.0 &	\;\;0.3 &	65.6 &	33.0	& 38.7\\
    ALBERT-base \cite{albert}& 32.2	& 43.1	& 37.7	& \;\;0.0	& 56.4	& 28.2 &	\;\;0.0	& 59.2	& 29.6	& 31.8 \\
    ALBERT-xxlarge & 32.1	&37.5	&34.8	&\;\;0.9	&50.5	&25.7	&\;\;0.3	&50.3	&25.3	&28.6\\
    RoBERTa-base \cite{roberta}&  28.6	&50.5	&39.5	&\;\;0.0	&46.1	&23.0&	\;\;0.1&	63.6&	31.8&	31.5\\
    RoBERTa-large & 34.2&	55.9&	45.0&	\;\;0.2&	50.6&	25.4	&\;\;0.9&	50.9&	25.9&	32.1\\
    XLNet-base \cite{xlnet}&  23.2	 & 49.1 & 	36.1 & 	\;\;1.9 & 	65.6 & 	33.8 & 	\;\;0.8 & 	63.5	 & 32.2 & 	34.0\\
    XLNet-large & 23.4&	42.0&	32.7&	\;\;4.7	&53.5&	29.1&	\;\;5.6&	48.4&	27.0&	29.6\\\midrule
    
    SBERT-base \cite{sbert}& 71.2	&73.7&	72.4&	41.8&	63.6&	52.7&	23.2&	58.7&	40.9&	55.3\\
    SBERT-large& 72.5	&74.2	&73.3&	57.8&	55.0&	\textbf{56.4}&	18.4&	52.4&	35.4	&55.0\\
    SALBERT-base& 75.8	&78.7	&77.3	&9.4&	63.6&	36.5&	3.8&	60.5&	32.1&	48.6\\
    SALBERT-xxlarge & 69.3	&77.0&	73.2&	55.8	&51.0&	53.4&	56.4&	48.8&	52.6&	59.7\\\midrule
    
    ALBURT-xxlarge & 68.7	& 82.1& 	75.4& 	54.2& 	56.1& 	55.1& 	54.2	& 52.0& 	\textbf{53.1}& 	\textbf{61.2}\\\bottomrule
  \end{tabular}
%   }
  \caption{Performance of different models on universal analogy datasets. ``sem": ``semantic", ``syn": ``syntactic". Mean-pooling is applied to Transformer-based models to obtain fixed-length embeddings. The last column shows the average accuracy of word, phrase and sentence analogy tasks. ALBURT-xxlarge represents our implementation.}
  \label{analogy_acc}
\end{table*}

\section{Evaluation}
We evaluate various models on our universal analogy datasets to investigate regularities of the learned vector space. 

\subsection{Embedding Methods}
The models we evaluate include Bag-of-words (BoW) model from pre-trained word embeddings, universal sentence embedding models, pre-trained contextualized language models, and fine-tuned contextualized representation models. In addition, we implement a Transformer-based model that learns universal representations from NLI and PPDB datasets.

\subsection{Our Method}
Given the effectiveness of the Siamese network in learning sentence embeddings \cite{sbert}, we introduce ALBURT, our implementation of  
multi-task training on ALBERT with the twin structure, to further enhance the universal representation. The training data includes NLI and the Paraphrase Database (PPDB) \cite{ppdb}. The former consists of the Stanford Natural Language Inference (SNLI) \cite{snli} and the Multi-Genre NLI Corpus \cite{multinli} that are frequently used to learn sentence representations. For the latter, we take advantage of the \textit{phrasal} PPDB dataset with S size, which contains 1.53 million multiword to single/multiword pairs. Relationships between pairs fall into six categories: \textit{Equivalence}, \textit{ForwardEntailment}, \textit{ReverseEntailment}, \textit{Independent}, \textit{Exclusion} and \textit{OtherRelated}. We apply a preprocessing step to the raw data:
\begin{enumerate}
    \item Pairs that are labeled with \textit{Exclusion} or \textit{OtherRelated} are filtered out.
    \item Examples of \textit{ForwardEntailment} and \textit{ReverseEntailment} are merged into one subset and relabeled as \textit{Entailment} since our model structure is symmetrical.
    \item We randomly select 343k pairs from each of the three labels: \textit{Equivalence}, \textit{Entailment} and \textit{Independent}, resulting in a total of 1.03 million examples.
\end{enumerate}

\subsubsection{Sentence-level Natural Language Inference}
Natural Language Inference (NLI) is a pairwise classification problem that is to identify the relationship between the premise and hypothesis from \textit{entailment}, \textit{contradiction}, and \textit{neutral}. It is considered as our sentence-level training objective.

\subsubsection{Phrase/word-level Paraphrase Identification}
Each pair in the PPDB dataset involves a target and its paraphrase. Using the negative sampling strategy \cite{phrase2vec}, we randomly sample $k$ ($k=3$ in our implementation) sequences for each target and annotate them with negative labels, indicating they are not paraphrases. Then the model is trained to distinguish between paraphrases and non-paraphrases.

\begin{figure}[t]
\centering
\includegraphics[width=0.4\textwidth]{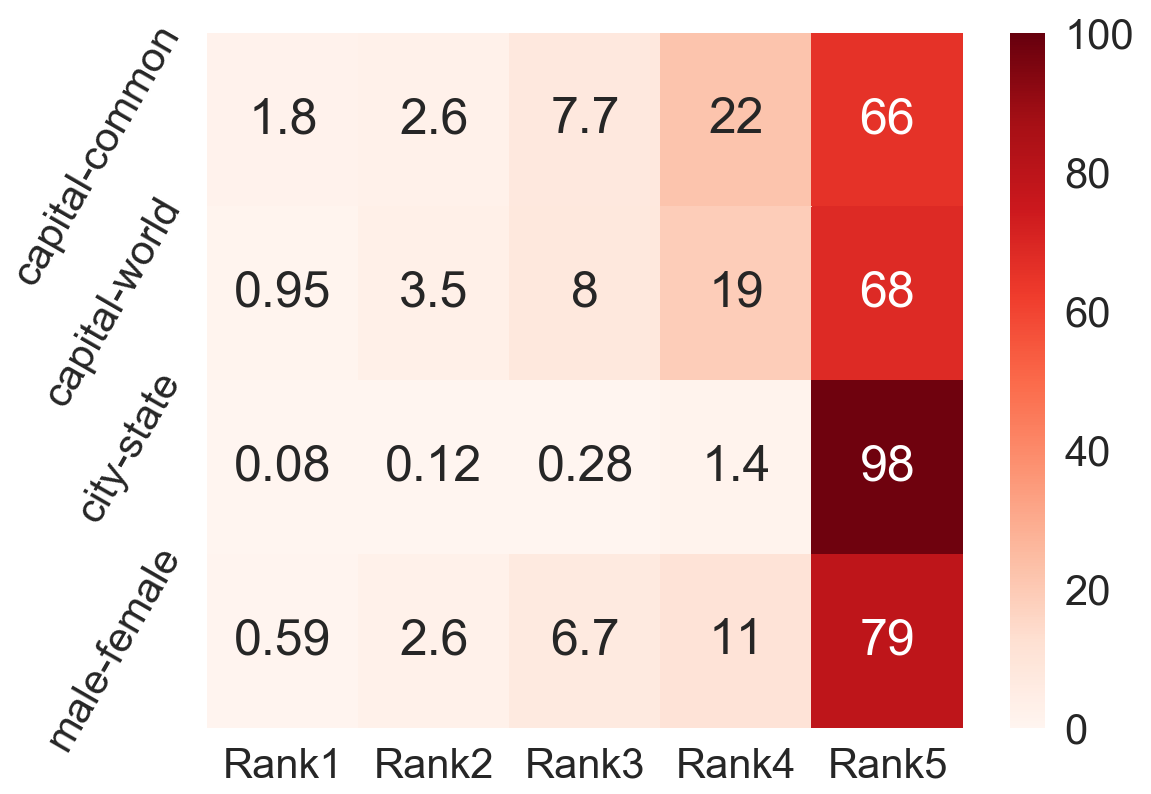}
\caption{Performance of USE on semantic datasets.}
\label{use}
\end{figure}

\begin{figure}[t]
\centering
\includegraphics[width=0.4\textwidth]{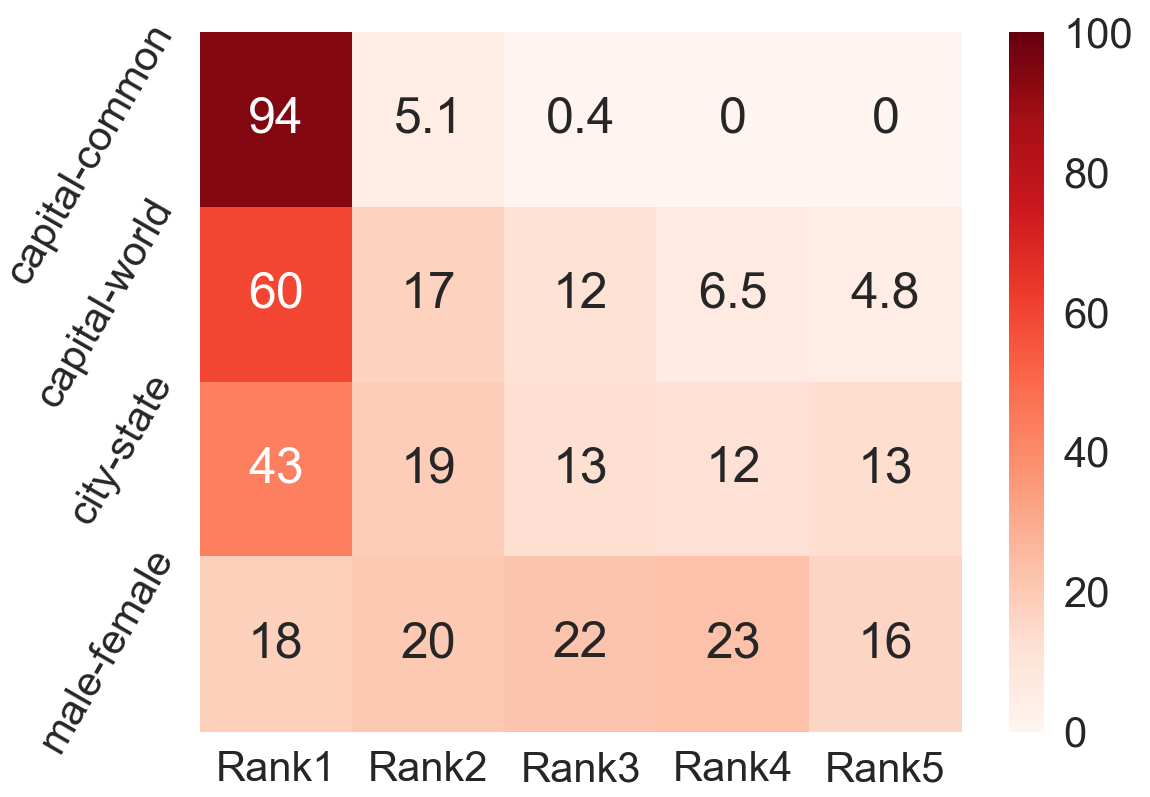}
\caption{Performance of ALBURT on semantic datasets.}
\label{alburt}
\end{figure}

\subsection{Phrase/word-level Entailment Classification}
Apart from the paraphrase identification task, we design a word/phrase-level entailment classification problem. For each paraphrase pair, the model is trained to recognize from three types of relationships: \textit{Equivalence}, \textit{Entailment} and \textit{Independent}. Thus, the model is forced to capture the degree of similarity between phrases and words while sequences are considered dissimilar even if they are closely related. 

\begin{table*}[t]
  \centering
  \small
%   \resizebox{\textwidth}{!}{
  \begin{tabular}{lc}\toprule
    \multicolumn{2}{l}{\textit{Barton's inquiry was reasonable }:\textit{ Barton's inquiry was not reasonable }::\textit{ Changing the sign of numbers is an efficient algorithm}}\\
    
    \textit{\textbf{changing the sign of numbers is an inefficient algorithm}}  & GloVe: 0.96\;\;\;\;SALBERT: \textbf{0.67}\;\;\;\;ALBURT: \textbf{0.73} \\
    \textit{changing the sign of numbers is not an inefficient algorithm} & GloVe: \textbf{0.97}\;\;\;\;SALBERT: 0.50\;\;\;\;ALBURT: 0.45 \\\midrule
    
    \multicolumn{2}{l}{\textit{Members are aware of their political work }:\textit{ Members are not aware of their political work }::\textit{ The associated ant is known}}\\
    
    \textit{\textbf{The associated ant is unknown}}  & GloVe: 0.94\;\;\;\;SALBERT: \textbf{0.79}\;\;\;\;ALBURT: \textbf{0.85}\\
    \textit{The associated ant is not unknown} & GloVe: \textbf{0.95}\;\;\;\;SALBERT: 0.63\;\;\;\;ALBURT: 0.77 \\\midrule
  \end{tabular}
%   }
  \caption{Questions and candidates from the sentence-level ``positive-negative" analogy dataset and similarity scores for each candidate sentence computed by GloVe, SALBERT and ALBURT. The correct sentences are in bold.}
  \label{pos-neg_example}
\end{table*}

\begin{table}[t]
  \centering
%   \resizebox{\textwidth}{!}{
  \renewcommand{\multirowsetup}{\centering}
  \begin{tabular}{l|cccc}\toprule
  & \multicolumn{2}{c}{Phrase} & \multicolumn{2}{c}{Sentence} \\\midrule
    Method & PPR & PNR     & PPR  & PNR     \\\midrule
    Glove  &\;\;0.0& \;\;0.0 &\;\;0.0& \;\;0.0	 \\
    USE    &\;\;1.7& \;\;0.0 & \;\;0.8  &\;\;0.0   \\
    SALBERT&\textbf{73.7}& \;\;3.4 &\textbf{74.0}&\;\;5.0   \\
    ALBURT &  72.7 &\textbf{10.1}&73.0&\textbf{13.0} \\\bottomrule
  \end{tabular}
%   }
  \caption{$PPR$ and $PNR$ on phrase and sentence datasets.}
  \label{metric}
\end{table}

\subsubsection{Multi-task Learning}
At each training step in the multi-task learning stage, a batch is randomly selected from the NLI dataset to train the model. Then we alternately choose one of the PPDB-related tasks (paraphrase identification/entailment classification) and randomly fetch a batch from the dataset. The models is trained for 2 epochs on 2 1080Ti GPUs, using Adam optimizer with learning rate 2e-5. Batch size is set to 16 and dropout rate is 0.1.

The lower layers are initialized with ALBERT and shared between the two inputs. For each task, two input sequences are tokenized and encoded separately and fed into a mean-pooling layer, resulting in two fixed-length vectors $u$ and $v$. We then compute $[u; v; |u-v|]$, which is the concatenation of the two representations and the absolute value of their difference, and finally feed it to a task-specific fully-connected layer followed by a softmax classification layer. 

\subsection{Results and Analysis}
Results on analogy tasks are reported in Table \ref{analogy_acc}. We can make the following conclusions.

(1) Generally, semantic analogies are more challenging than the syntactic ones and higher-level relationships between sequences are more difficult to capture. This phenomenon is observed in almost all the evaluated models.

(2) On word analogy tasks, USE achieves the highest accuracy (85.1\%). All well pre-trained language models like BERT, ALBERT, RoBERTa and XLNet hardly exhibit arithmetic characteristics and increasing the model size usually leads to a decrease in accuracy. However, properly fine-tuning BERT and ALBERT on labeled datasets greatly improves the model performance and get the accuracy of 73.3\% (SBERT) and 75.4\% (ALBURT), respectively. 

(3) Despite the leading performance on word-level analogy datasets of GloVe, InferSent and USE, they do not generalize well on higher level analogy tasks. In contrast, Transformer-based models are more advantageous in representing higher-level sequences. Overall, ALBURT achieves the highest average accuracy,  which shows that it has indeed learned universal representations across different linguistic units.

(4) In order to compare the performance of USE and ALBURT on different categories of semantic analogies, we analyse the proportion of rank values given to the correct sentences by both models in Figure \ref{use} and Figure \ref{alburt}. We conjecture that the poor performance of USE is caused by synonyms and paraphrases in sentences which lead USE to produce lower similarity scores to the correct answers. However, trained on massive NLI and paraphrase data, ALBURT is especially good at identifying paraphrases and capturing relationships between sentences even if they have less lexical overlapping.

(5) Examples from the Negation subset are shown in Table \ref{pos-neg_example}. Notice that the word ``\textit{not}" does not explicitly appear in the correct answers. Instead, ``\textit{inefficient}" and ``\textit{unaware}" are indicators of negation. As expected, BOW will give a higher similarity score for the sentence that contain both ``\textit{not}" and ``\textit{inefficient}" because the word-level information is simply added and subtracted despite the context. By contrast, contextualized models like SALBERT and ALBURT capture the meanings and relationships of words within the sequence in a comprehensive way.

\subsection{Further Analysis}
Besides the discussion of results on analogy tasks with respect to each linguistic level, we are also interested in the relationship of model performance on different datasets. Since our phrase and sentence-level semantic datasets are developed from the word-level questions, we inspect model generalization from words to higher level sequences. Thus, we introduce two metrics: $PPR$ and $PNR$:
\begin{align*}
    & PPR = \frac{|\mathcal{PP}|}{|\mathcal{P}|}
    & PNR = \frac{|\mathcal{PN}|}{|\mathcal{N}|}
\end{align*}
where $\mathcal{P}$ and $\mathcal{N}$ are the set of questions that are answered correctly and incorrectly in the word dataset, respectively. $\mathcal{PP}$ and $\mathcal{PN}$ are subsets of $\mathcal{P}$ and $\mathcal{N}$, respectively. In addition, questions in $\mathcal{PP}$ and $\mathcal{PN}$ are answered correctly in the phrase/sentence dataset.

Specifically, $PPR$ examines the proportion of questions which are correctly answered in the word dataset that can still be solved in higher-level analogies, and $PNR$ explores the ratio of examples where the model fail in word analogy that can be corrected in phrase/sentence analogies. Therefore, we want both $PPR$ and $PNR$ to be high. 

According to the results in Table \ref{metric}, SALBERT and ALBURT substantially outperform USE and GloVe on phrase and sentence-level $PPR$, which means most of the relationships they identify in word questions remain to be answered correctly in phrase and sentence datasets. Surprisingly, ALBURT achieves 10.1 and 13.0 of $PNR$ on phrase and sentence datasets, respectively, which is considerable compared with GloVe (0.0/0.0), USE (0.0/0.0) and SALBERT (3.4/5.0). This can be explained that words like ``\textit{apple}" and ``\textit{china}" have multiple meanings and often appear in certain contexts. Manipulating their embeddings in isolation will cause contextualized representation models confused. However, phrases such as ``\textit{eat an apple}" and ``\textit{university in China}" help ALBURT understand the contextual meaning of each word. Thus, ALBURT is able to recover the relationship between sequences even though it fails on the word-level questions.

\begin{table}[t]
%   \centering
  \small
%   \resizebox{\textwidth}{!}{
  \begin{tabular}{|ll|}\hline
   
    p\_man: \textit{employed by the \underline{man}} & p\_woman: \textit{hired by the \underline{woman}}\\
    
    p\_king: \textit{employed by the \underline{king}} & p\_queen: \textit{hired by the \underline{queen}}\\
   
  p\_dad: \textit{employed by his \underline{dad}} & p\_mom: \textit{hired by his \underline{mom}}\\\hline
   
  \multicolumn{2}{|l|}{s\_man: \textit{He was employed by the \underline{man} when he was 28.}} \\
    \multicolumn{2}{|l|}{s\_woman: \textit{He was hired by the \underline{woman} at age 28.}} \\
    \multicolumn{2}{|l|}{s\_king: \textit{He was employed by the \underline{king} when he was 28.}} \\
    \multicolumn{2}{|l|}{s\_queen: \textit{He was hired by the \underline{queen} at age 28.}} \\
    \multicolumn{2}{|l|}{s\_dad: \textit{He was employed by his \underline{dad} when he was 28.}} \\
  \multicolumn{2}{|l|}{s\_mom: \textit{He was hired by his \underline{mom} at age 28.}} \\\hline

  \end{tabular}
%   }
  \caption{Annotation of phrases and sentences in Figure \ref{distance}.}
  \label{explanation}
\end{table}

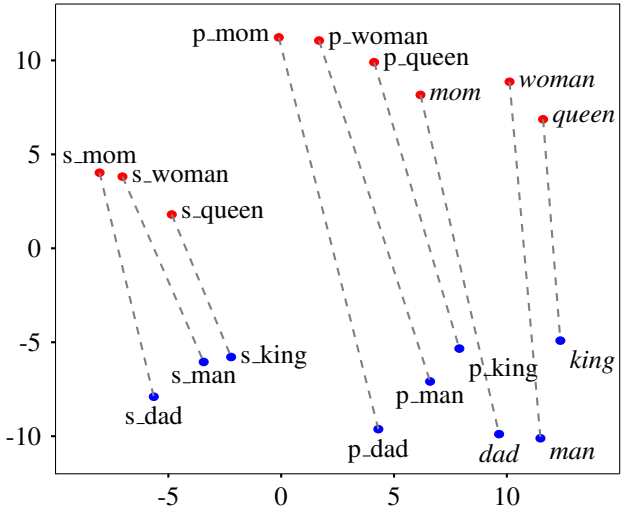
\begin{figure}[t]
\centering
\begin{tikzpicture} [xscale=0.3, yscale=0.25]
\draw [-] (-10,13) -- (-10,-12) -- (15,-12);
\draw [-] (-10,13) -- (15,13) -- (15,-12);
% \draw [thick] (-10,-12.2) node[below]{-10} -- (-10,-12);
\draw [thick] (-5,-12.2) node[below]{-5} -- (-5,-12);
\draw [thick] (0,-12.2) node[below]{0} -- (0,-12);
\draw [thick] (5,-12.2) node[below]{5} -- (5,-12);
\draw [thick] (10,-12.2) node[below]{10} -- (10,-12);
% \draw [thick] (15,-12.2) node[below]{15} -- (15,-12);
\draw [thick] (-10.2,-10) node[left]{-10} -- (-10,-10);
\draw [thick] (-10.2,-5) node[left]{-5} -- (-10,-5);
\draw [thick] (-10.2,0) node[left]{0} -- (-10,0);
\draw [thick] (-10.2,5) node[left]{5} -- (-10,5);
\draw [thick] (-10.2,10) node[left]{10} -- (-10,10);
\draw [fill,blue] (12.370384, -4.9123874) circle [radius=0.2];
\draw [fill,blue] (9.656271, -9.889573) circle [radius=0.2];
\draw [fill,blue] (11.487694, -10.109503) circle [radius=0.2];
\draw [fill,blue] (7.8951035, -5.3361106) circle [radius=0.2];
\draw [fill,blue] (4.3060765, -9.624377) circle [radius=0.2];
\draw [fill,blue] (6.5985246, -7.0880723) circle [radius=0.2];
\draw [fill,blue] (-2.21838, -5.786036) circle [radius=0.2];
\draw [fill,blue] (-5.64602, -7.897916) circle [radius=0.2];
\draw [fill,blue] (-3.434485, -6.0478945) circle [radius=0.2];
\draw [fill,red] (11.604825, 6.864548) circle [radius=0.2];
\draw [fill,red] (6.172143, 8.166092) circle [radius=0.2];
\draw [fill,red] (10.11775, 8.862421) circle [radius=0.2];
\draw [fill,red] (4.1176953, 9.903919) circle [radius=0.2];
\draw [fill,red] (-0.10769433, 11.222963) circle [radius=0.2];
\draw [fill,red] (1.6783793, 11.057503) circle [radius=0.2];
\draw [fill,red] (-4.850041, 1.7995701) circle [radius=0.2];
\draw [fill,red] (-8.048725, 4.023928) circle [radius=0.2];
\draw [fill,red] (-7.033974, 3.8145025) circle [radius=0.2];
\draw [dashed, gray, thick] (-8.048725  , 4.023928) -- (-5.64602,   -7.897916);
\draw [dashed, gray, thick] (-7.033974 ,  3.8145025) -- (-3.434485,  -6.0478945);
\draw [dashed, gray, thick] (-4.850041, 1.7995701) -- (-2.21838,  -5.786036);
\draw [dashed, gray, thick] (-0.10769433 ,11.222963) -- (4.3060765, -9.624377);
\draw [dashed, gray, thick] (1.6783793 ,11.057503) -- (6.5985246, -7.0880723);
\draw [dashed, gray, thick] (4.1176953 ,9.903919) -- (7.8951035, -5.3361106);
\draw [dashed, gray, thick] (6.172143, 8.166092) -- (9.656271, -9.889573);
\draw [dashed, gray, thick] (10.11775,   8.862421) -- (11.487694, -10.109503);
\draw [dashed, gray, thick] (11.604825 , 6.864548) -- (12.370384,  -4.9123874);
\node [right] at (-7.033974 ,  3.8145025) {s\_woman};
\node [right] at (-4.850041, 1.7995701) {s\_queen};
\node [left] at (-0.10769433 ,11.222963) {p\_mom};
\node [right] at (1.6783793 ,11.057503) {p\_woman};
\node [right] at (4.1176953 ,9.903919) {p\_queen};
\node [right, font=\itshape] at (6.172143, 8.166092) {mom};
\node [right, font=\itshape] at (10.11775,   8.862421) {woman};
\node [right, font=\itshape] at (11.604825 , 6.864548) {queen};
\node [above] at (-8.048725,4.023928) {s\_mom};

\node [below] at (-3.434485,  -6.0478945) {s\_man};
\node [right] at (-2.21838,  -5.786036) {s\_king};
\node [below] at (4.3060765, -9.624377) {p\_dad};
\node [below] at (6.5985246, -7.0880723) {p\_man};
\node [below right] at (7.8951035, -5.3361106) {p\_king};
\node [below, font=\itshape] at (9.656271, -9.889573) {dad};
\node [below right, font=\itshape] at (11.487694, -10.109503) {man};
\node [below right,font=\itshape] at (12.370384,  -4.9123874) {king};
\node [below] at (-5.64602,   -7.8979168) {s\_dad};
\end{tikzpicture}
\caption{Two-dimensional PCA projection of the vectors representing ``male" and ``female" generated by our ALBURT model. Pairs are connected by dashed lines. Each point in the figure represents a word, phrase or sentence, as explained in detail in Table \ref{explanation}.}
\label{distance}
\end{figure}

\begin{figure}[t]
\centering
\includegraphics[width=0.5\textwidth]{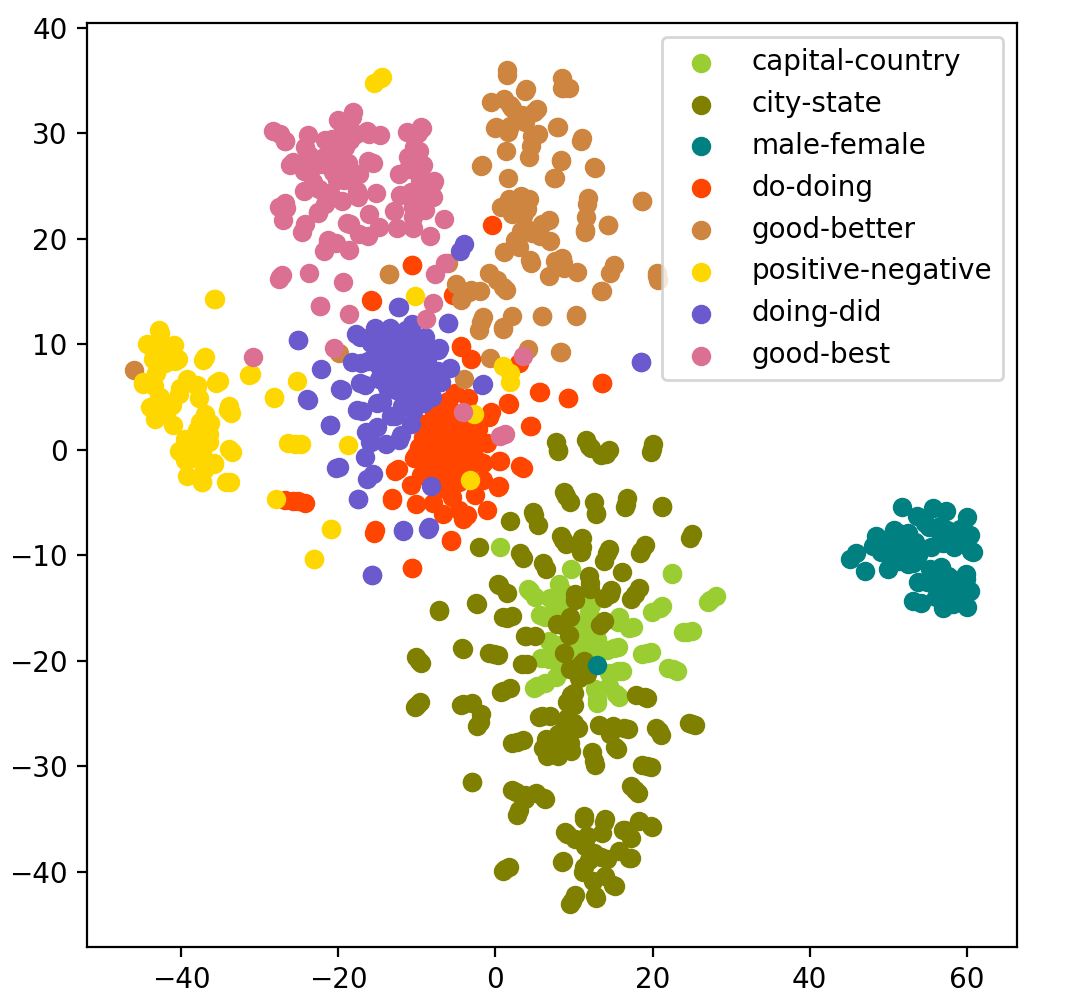}
\caption{t-SNE projection of patterns.}
\label{cluster}
\end{figure}

\subsection{Visualization}
\subsubsection{Single Pattern}
\cite{phrase2vec} use PCA to project word embeddings into a two-dimensional space to visualize a single pattern captured by the model, while in this work we consider embeddings for different granular linguistic units. All pairs in Figure \ref{distance} belong to the ``male-female" category and subtracting the two vectors results in roughly the same direction. We also notice that sentences are closer to each other compared with phrase and word pairs, which is reasonable because longer sequences have more lexical overlapping.

\subsubsection{Clustering}
Given that embeddings of sequences with the same kind of relationship will exhibit the same pattern in the vector space, we obtain the difference between pairs of embeddings for words, phrases and sentences from different categories and visualize them by t-SNE. Figure \ref{cluster} shows that by subtracting two vectors, pairs that belong to the same category automatically fall into the same cluster. Only the pairs from ``capital-country" and ``city-state" cannot be totally distinguished because they all describe the relationship between geographical entities.

\section{Application on FAQ}
Transformer-based universal representation models that are fine-tuned on a large amount of data have shown great performance on our analogy tasks. We can safely draw two conclusions: semantic and syntactic relationships between sequences can be reflected by simple arithmetic operations in the same vector space; distance between sequences of different lengths can be easily measured by cosine similarity using the pre-computed embeddings. Based on these observations, we present an insurance FAQ dataset and apply universal representation models on it to study their effect in real-world applications.

\subsection{Dataset}
We collect real-world frequently asked questions and answers between users and customer service from our partners in a Chinese online financial education institution. It contains over 4 types of insurance questions (e.g. concept explanation (``\textit{what}"), insurance consultation (``\textit{why}", ``\textit{how}"), judgement (``\textit{whether}") and recommendation. The dataset is composed of 300 Question-Answer pairs that are carefully selected to avoid similar questions so that each query has only one exact match. The training and test sets consist of 1200 and 200 queries, respectively. 

\subsection{Task Description}
In this work, we focus on the insurance FAQ retrieval task. Our collection of QA pairs can be denoted as $\{(Q_1, A_1), (Q_2, A_2), ... (Q_N, A_N)\}$, where $N$ is the number of QA pairs. For each query, the goal is to retrieve an appropriate QA pair from the pre-defined dataset. The query-Question similarity is regarded as a measure of ranking in this task.

\begin{figure*}[t]
\centering
\begin{tcolorbox}[left = 1mm, right = 0mm, top = 0mm, bottom = 0mm, boxsep = 0mm,
      toptitle = 1mm, bottomtitle = 1mm,colbacktitle=black!10!white,colback=red!0!white,coltitle=black!100!white, title = {\textit{ Can 80-year-old people get accident insurance? \qquad\quad Is there any insurance that you recommend?}}]
{\arrayrulecolor{gray}
    \renewcommand{\arraystretch}{1.2}
    \centering
    \begin{tabular}{p{1.1cm}|p{5.8cm}|p{9.2cm}}
      BM25   & \textit{Can I get insurance after an accident?} & \textit{Will managers recommend products to clients for their own benefit?} \\\hline
      TF-IDF & \textit{Can life insurance last until the age of 80?} & \textit{Can I get insurance for my boyfriend?} \\\hline
      XLNet &\textit{Can seniors buy accident insurance? }\CheckmarkBold & \textit{Which insurance is suitable for me? }\CheckmarkBold
      \end{tabular}}\hfill
\end{tcolorbox}
\caption{Examples of queries and responses.}
\label{faq_example}
\end{figure*}

\begin{table}[t]
  \centering
%   \small
%   \resizebox{\textwidth}{!}{
  \renewcommand{\multirowsetup}{\centering}
  \begin{tabular}{l|rrr}\arrayrulecolor{black} \toprule
    Method  & Acc. & Std. &MRR\\\midrule
    TF-IDF  &	82.5 &- & 0.875\\
    BM25	&82.5 & - & 0.856\\\midrule
    BERT-base&	88.0&- &0.917 \\
    XLNet-base&	74.5 &- &0.805\\
    XLNet-mid 	&46.0 & -&0.534 \\\midrule
    BERT-base + NLI	&83.0 &-&0.875  \\
    XLNet-base + NLI&	83.5   &-&0.879 \\
    XLNet-mid + NLI&	84.5    &- &0.881\\\midrule
    BERT-base + FAQ &94.4 &  2.2  &0.961  \\
    XLNet-base + FAQ& 78.2 &18.1    &0.824 \\
    XLNet-mid + FAQ& 82.3    & 22.8 &0.848 \\\midrule
    BERT-base + NLI + FAQ&94.3  & 2.1&0.948\\
    XLNet-base + NLI + FAQ & 94.2 &  1.4&0.952\\
    XLNet-mid + NLI + FAQ  &   \textbf{96.2} & \textbf{1.2}&\textbf{0.974}\\\bottomrule
  \end{tabular}
%   }
  \caption{Comparison of statistical methods and contextualized language models on the FAQ dataset.}
  \label{faq_acc}
\end{table}

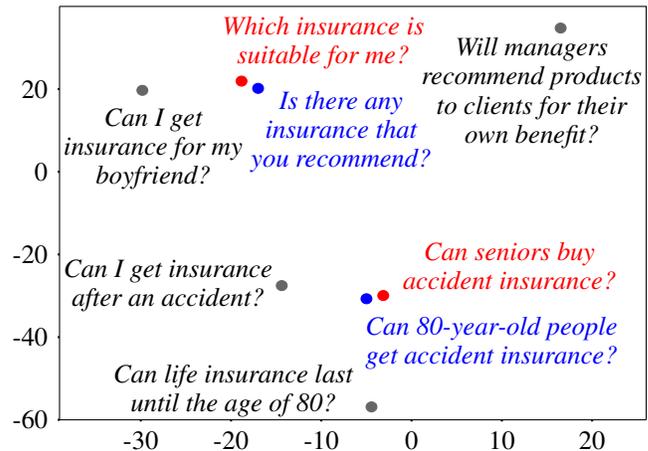
\begin{figure}[t]
\centering
\begin{tikzpicture} [xscale=0.12, yscale=0.11]
\draw [-] (-39,20) -- (-39,-30) -- (26,-30);
\draw [-] (-39,20) -- (26,20) -- (26,-30);
\draw [thick] (-30,-30.2) node[below]{-30} -- (-30,-30);
\draw [thick] (-20,-30.2) node[below]{-20} -- (-20,-30);
\draw [thick] (-10,-30.2) node[below]{-10} -- (-10,-30);
\draw [thick] (0,-30.2) node[below]{0} -- (0,-30);
\draw [thick] (10,-30.2) node[below]{10} -- (10,-30);
\draw [thick] (20,-30.2) node[below]{20} -- (20,-30);
\draw [thick] (-39.2,-30) node[left]{-60} -- (-39,-30);
\draw [thick] (-39.2,-20) node[left]{-40} -- (-39,-20);
\draw [thick] (-39.2,-10) node[left]{-20} -- (-39,-10);
\draw [thick] (-39.2,0) node[left]{0} -- (-39,0);
\draw [thick] (-39.2,10) node[left]{20} -- (-39,10);
\draw [fill,blue] (-17.002241,  10.090749) circle [radius=0.6];
\draw [fill,red] (-18.837675 , 10.9661675) circle [radius=0.6];
\draw [fill,blue] (-4.9968176 ,-15.3560965) circle [radius=0.6];
\draw [fill,red] (-3.138091 ,-14.9648125) circle [radius=0.6];
\draw [fill,black!60] (16.518122 ,17.3846875) circle [radius=0.6];
\draw [fill,black!60] (-29.824173  ,9.8515625) circle [radius=0.6];
\draw [fill,black!60] (-4.438696 ,-28.43617) circle [radius=0.6];
\draw [fill,black!60] (-14.381388 ,-13.763403) circle [radius=0.6];
\node [above right,align=center,red, font=\itshape] at (-22.002241,  11.290749) {Which insurance is\\suitable for me?};
\node [below right,align=center,blue, font=\itshape] at (-18.837675 , 10.9661675) {Is there any\\insurance that\\you recommend?};
\node [above right,align=center,red, font=\itshape] at (-1.9968176 ,-15.3560965) {Can seniors buy\\accident insurance?};
\node [below right,align=center,blue, font=\itshape] at (-6.138091 ,-16.3648125) {Can 80-year-old people\\get accident insurance?};
\node [below,align=center,black, font=\itshape] at (-28.624173  ,8.8515625) {Can I get\\insurance for my\\boyfriend?};
\node [left,align=center,black, font=\itshape] at (-14.381388 ,-13.763403) {Can I get insurance\\after an accident?};
\node [left,align=center,black, font=\itshape] at (-5.438696 ,-26.43617) {Can life insurance last\\until the age of 80?};
\node [below,align=center,black, font=\itshape] at (13.318122 ,17.3846875) {Will managers\\recommend products\\to clients for their\\own benefit?};
\end{tikzpicture}
\caption{t-SNE projection of XLNet embeddings.}
\label{faq_vector}
\end{figure}

\subsection{Experiments and Results}
We create negative training data through a retrieval-based negative sampling strategy. Specifically, for each query in the training set, we mark the top 4 QA pairs (except for the correct answer) retrieved by the pre-trained BERT model with negative labels. We find that this retrieval-based sampling method enables the model to obtain an improvement of $\sim$ 3\% accuracy compared with random sampling.

We evaluate Transformer-based models and statistical methods such as IF-IDF and BM25 on the FAQ task. Structure of universal representation models follows the Siamese network as described in the last section and pre-trained BERT and XLNet are used for initialization. Our evaluation is based on Top-1 Accuracy (Acc.) and Mean Reciprocal Rank (MRR). Models are trained with 5 random seeds for fine-tuning on the FAQ dataset and average accuracy as well as standard deviation on the test set are reported in Table \ref{faq_acc}. 

BERT and XLNet models outperform IF-IDF and BM25 significantly. Even the pre-trained BERT model achieves an accuracy of 88\%, and fine-tuning BERT on the FAQ dataset gives 6.4\% improvement. Fine-tuning on NLI datasets has a large impact on XLNet, but hurts the performance of BERT. We also find that the NLI dataset helps lower the deviation on the test set. By first training on the NLI data, then fine-tuning on the FAQ dataset, XLNet-mid obtains the highest accuracy and MRR and the lowest standard deviation. We conclude that the NLI dataset enables the model to capture the meaning of a sentence, and training on the domain-specific FAQ dataset help the model adapt to the target data.

We show examples in Table \ref{faq_example} where the fine-tuned XLNet model successfully retrieve the correct answer while TF-IDF and BM25 fail. Both sentences ``\textit{Can 80-year-old people get accident insurance?}" and ``\textit{Can life insurance last until the age of 80?}" contain the word ``\textit{80}", which is a possible reason why TF-IDF tends to believe they highly match with each other, ignoring that the two sentences are actually describing two different issues. In contrast, using vector-based representations, XLNet considers ``\textit{seniors}" as a paraphrase of ``\textit{80-year-old people}". As depicted in Figure \ref{faq_vector}, queries are close to the correct responses and away from other sentences.

\section{Conclusion}
This work concentrates on the less concentrated language representation, seeking to learn a uniform vector form across different linguistic unit hierarchies. Far apart from learning either word only or sentence only representation, we find that training Transformer models on a large-scale corpus effectively learns a universal representation from words, phrases to sentences. We especially provide universal analogy datasets 
\footnote{Our annotated datasets will be publicly released after the anonymous reviewing period.} 
and an insurance FAQ dataset to evaluate models from different perspectives. The well-trained universal representation model holds the promise for demonstrating accurate vector arithmetic with regard to words, phrases and sentences and in applications such as FAQ retrieval tasks.

\bibliography{aaai21}
\bibliographystyle{aaai21}

\end{document}